\documentclass[conference]{IEEEtran}
\IEEEoverridecommandlockouts

\usepackage{todonotes}
\usepackage{url}
\usepackage[table,xcdraw]{xcolor}
\usepackage{booktabs}
\usepackage{multirow}
\usepackage{cite}
\usepackage{amsmath,amssymb,amsfonts}
\usepackage{algorithmic}
\usepackage{graphicx}
\usepackage{textcomp}
\def\BibTeX{{\rm B\kern-.05em{\sc i\kern-.025em b}\kern-.08em
    T\kern-.1667em\lower.7ex\hbox{E}\kern-.125emX}}
\usepackage{hyperref}
\hypersetup{
    colorlinks=true,
    allcolors=blue,
    pdftitle={Network Traffic Analysis with Process Mining: The UPSIDE Case Study},
    pdfpagemode=FullScreen,
}

\begin{document}

\title{Network Traffic Analysis with\\ Process Mining: The UPSIDE Case Study
\thanks{This work was supported by the Spoke 9 “Digital Society \& Smart Cities” of ICSC - Centro Nazionale di Ricerca in High Performance-Computing, Big Data and Quantum Computing, funded by the European Union - NextGenerationEU (PNRR-HPC, CUP: E63C22000980007), and the UPSIDE project (B63D23000820004), funded by the Italian MUR.}
}

\author{
\IEEEauthorblockN{Francesco Vitale, Massimiliano Rak, Nicola Mazzocca}
\IEEEauthorblockA{
\textit{Department of Electrical Engineering and Information Technology}\\
\textit{University of Naples Federico II}\\
Naples, Italy\\
\{francesco.vitale, massimiliano.rak, nicola.mazzocca\}@unina.it}
\and
\IEEEauthorblockN{Paolo Palmiero}
\IEEEauthorblockA{
\textit{IMT School for Advanced Studies Lucca}\\
Lucca, Italy\\
paolo.palmiero@imtlucca.it}
}

\maketitle

\begin{abstract}
Online gaming is a popular activity involving the adoption of complex systems and network infrastructures. The relevance of gaming, which generates large amounts of market revenue, drove research in modeling network devices' behavior to evaluate bandwidth consumption, predict and sustain high loads, and detect malicious activity. In this context, process mining appears promising due to its ability to combine data-driven analyses with model-based insights. In this paper, we propose a process mining-based method that analyzes gaming network traffic, allowing: unsupervised characterization of different states from gaming network data; encoding such states through process mining into interpretable Petri nets; and classification of gaming network traffic data to identify different video games being played. We apply the method to the UPSIDE case study, involving gaming network data of several devices interacting with two video games: Clash Royale and Rocket League. Results demonstrate that the gaming network behavior can be effectively and interpretably modeled through states represented as Petri nets with sufficient coherence and specificity while maintaining a good classification accuracy of the two different video games.
\end{abstract}


\section{Introduction}
\label{sec:intro}
The large landscape of digital applications spans several domains and includes many activities involving multiple interacting users. E-games are becoming an increasingly popular class of such applications, whose market revenue accounts for tens of billions of dollars across different platforms and paradigms, including cloud and mobile gaming \cite{baek2023cloudgaming}. The large interest in gaming makes analyzing the Internet traffic generated by this cyber application class relevant to evaluate network bandwidth consumption, predict and sustain high network loads, and detect malicious activity \cite{carrascosa2022cloudgaminganalysis}. 

While deep learning has been widely explored for Internet traffic analysis, such approaches often suffer from limited interpretability, motivating the integration of explainable techniques \cite{nascita2024explainableai}. In this context, process mining has also been proposed due to its ability to combine data-driven analyses with model-based insights, providing an explainable process-based view closely aligned to the actual behavior of the target system \cite{aalst2022pmhb, vitale2025pmdt}. In fact, many proposals outlined the explainable nature of process mining when analyzing the typical behavior of network protocols, including those widely employed in IoT applications such as MQTT and OPC UA \cite{ahmadon2020cpsmqttpbad, empl2024processawareidmqtt, hornsteiner2024pmopcua}. In particular, process mining can extract models that combine multi-perspective behavioral aspects, including packet-level traffic sequences and their impact on network load, including payload sizes and transmission times. This is particularly relevant in instrumentation and measurement, as it enables the systematic analysis of IoT communication behavior under realistic operating conditions, where resource constraints, periodic sensing, and protocol overhead strongly influence traffic patterns and device performance \cite{morato2021measurement}

However, despite the opportunities opened by process mining in explainable network traffic analysis, the nature of network traffic hinders its use. First, the noisy and interleaved nature of network traffic data leads to challenges in identifying meaningful events to enable the application of process mining algorithms \cite{bouhidel2023adhocnetworkspmanalysis, engelberg2023uncertainty}. Second, this complexity may lead to underfitting models that provide shallow generalizations of network traffic behavior \cite{dealvarenga2018pmhc}. Third, traffic data is often captured without prior knowledge of the activities that drove its generation \cite{hornsteiner2024pmopcua}. Besides, the literature lacks the application of process mining for network traffic analysis in video games. 

To address the above-mentioned challenges and literature gap, we propose a process mining-based method that encodes gaming network traffic states into behavioral models. The main novelty of our approach lies in:
\begin{itemize}
\item{unsupervised identification of different states in gaming network traffic;}
\item{fine-tuning of state space characterization with different complexity degrees to account for the noisy and interleaved nature of network data;}
\item{encoding the different states through process mining into interpretable behavioral models;}
\item{classification of gaming network data to identify different network states and detect the games being played.}
\end{itemize}

We applied our method to the UPSIDE case study, where gaming network data were monitored from several devices playing different online video games. Results outlined that our method was able to: 1) encode network traffic into different behavioral models that were, on the one hand, coherent across the devices, and, on the other hand, different from each other, and 2) classify the network traffic of different games being played.

The rest of the paper is organized as follows. Section \ref{sec:sota} reviews the use of process mining for network traffic analysis and the gaming datasets available in the literature. Section \ref{sec:method} describes the different phases of our method. Section \ref{sec:experiments} reviews the case study and our experiments. Section \ref{sec:conclusions} draws the conclusions and reviews future work.
\section{State-Of-The-Art}
\label{sec:sota}
In this section, we first review the literature on the state-of-the-art of process mining for network traffic analysis. Next, we review the available gaming datasets in the literature.

\subsection{Process Mining for Network Traffic Analysis}
Network traffic analysis includes modeling network protocols behavior, predicting network usage, and verifying deviations from expected behavior \cite{nascita2024explainableai}. Process mining enables these tasks by process discovery, conformance checking and process enhancement, which deal with the automatic discovery of interpretable process models, checking new behavior against such models, and adding other perspectives such as time and resource usage \cite{aalst2022pmhb, macak2022pmcybersecurity}. The literature identified these opportunities and offered several process mining-based solutions for network traffic analysis.

Saint-Pierre et al. \cite{saintpierre2014dnspmad} put forward a process-based approach employing process mining for modeling the DNS protocol and inspecting users' behavior. They claimed this approach could be useful for detecting network disruptions due to malicious activity. Ahmadon et al. \cite{ahmadon2020cpsmqttpbad} and Empl et al. \cite{empl2024processawareidmqtt} employed a similar strategy against the MQTT protocol, outlining the process-based, explainable benefits of employing process mining. The authors focused on the opportunities of their approach related to anomaly detection in cyber-physical systems. In addition, they show impressive model quality results, which outline the utility of process mining in accurately capturing the overall protocol behavior. Bouhidel and Belala \cite{bouhidel2023adhocnetworkspmanalysis} investigated the utility of process mining for modeling ad-hoc networks, self-organizing collections of mobile nodes that operate cooperatively. The authors aimed at collecting send/receive messages related to low-level network protocols, such as MAC, RTR and AGT protocols, and modeling the overall network behavior through process mining. Hadad et al. \cite{hadad2023networktrafficdata} proposed reconstructing business-level processes by inspecting network data flowing in response to specific activities being carried out by the business process of information systems. Zhong and Lisitsa \cite{zhong2022pmhelp} attempted to deal with the enormous amounts of data flowing across IoT networks, which can be vulnerable to, e.g., brute-force, DoS and botnet attacks. Specifically, they dealt with TCP data and showed the results of modeling the protocol behavior using process mining. Blefari et al. \cite{blefari2024logbasedad} proposed merging network and OS-level logs to discover attack paths in cyber range platforms. They showed that the behavior captured in the presence of attacks deviates significantly from normal behavior, which can help build attack profiles and recognize specific malicious behavior. Hornsteiner et al. \cite{hornsteiner2024pmopcua} investigated the utility of process mining against OPC UA, a widely used protocol in the industrial IoT.

Despite the aforementioned opportunities regarding the use of process mining for various tasks within network traffic analysis, many challenges remain. Specifically, although a few works showed impressive modeling results for various network protocols, including those widely used in the challenging IoT scenarios such as MQTT and OPC UA \cite{ahmadon2020cpsmqttpbad, empl2024processawareidmqtt, hornsteiner2024pmopcua}, the datasets used involved the controlled generation of network traffic, which is often not the case in the majority of monitored network data. In addition, several works have outlined the difficulties of identifying the so-called case IDs, namely different network flow instances within the noisy and interleaved network traffic \cite{saintpierre2014dnspmad, engelberg2023uncertainty, bouhidel2023adhocnetworkspmanalysis, hadad2023networktrafficdata, hornsteiner2024pmopcua}. Finally, the complexity of network traffic data also led to negative results, which claim the inadequacy of process mining to effectively capture network traffic behavior \cite{zhong2022pmhelp}. To address these issues, we propose an unsupervised method for modeling network traffic data that systematically addresses both the identification of case IDs and the management of the complexity of network traffic data.

\subsection{Gaming Datasets}
The public availability of gaming traffic datasets is scarce, especially those focused on large-scale competitive events \cite{upsideAina2025}. 
While there are datasets on general network traffic analysis, these are more inclined to explore other aspects of networking, lacking the particular focus on the gaming aspects. An example of such datasets is presented in reference \cite{labayenNetworkTrafficCode2020}, which only focuses on providing a collection of network traffic for classification purposes, but it does not consider at all any kind of gaming aspect that may exist on the network.
The same applies to the dataset found in reference \cite{bastianACIIoTNetwork2023}, which presents a particular dataset more inclined to show network utilization from IoT devices that have been placed and monitored in a controlled environment and later their traffic has been collected for machine learning model training purposes.
\begin{figure*}[!t]
\centering
\includegraphics[width=0.9\textwidth]{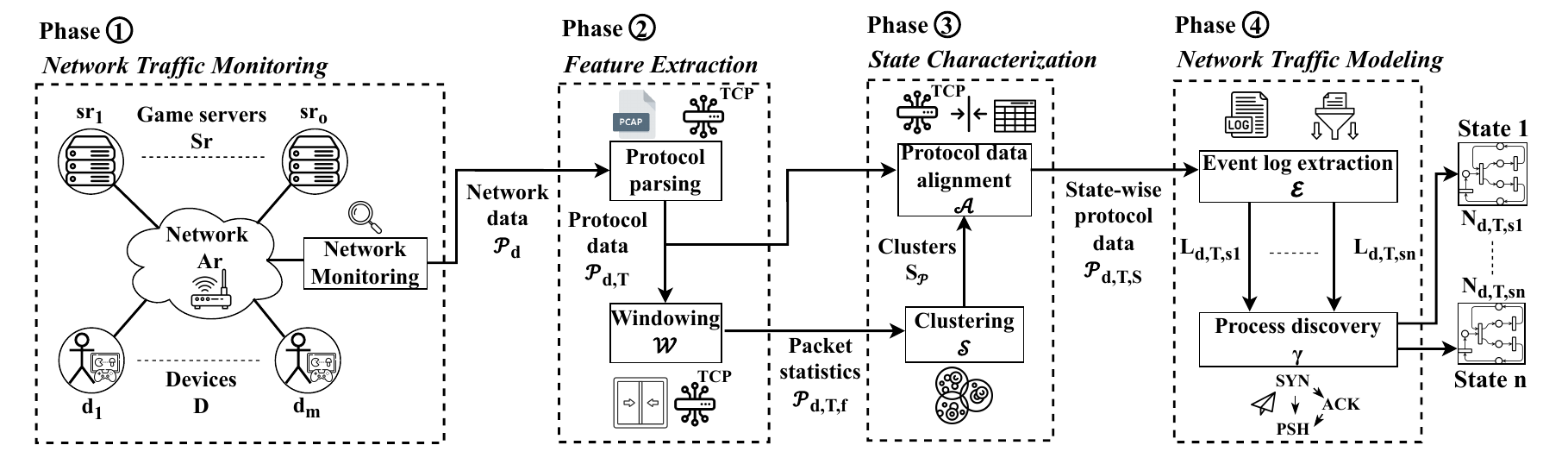}
\caption{The proposed method for encoding gaming traffic of network devices into behavioral models.}
\label{fig:method}
\end{figure*}

While there are a few datasets that focus specifically on gaming, they tend to examine individual gaming sessions or isolate particular game genres or titles. As a result, they fail to capture the dense, simultaneous traffic patterns characteristic of full-scale gaming events. This gap is significant, as the complexity of such environments is essential for accurate modelling —- something that synthetic data or isolated gameplay traffic cannot fully replicate, even with the aid of advanced machine learning techniques.

\section{The Proposed Method}
\label{sec:method}

To advance beyond standard traffic classification, this section introduces a method, depicted in Figure \ref{fig:method}, that aims to model the different states traversed by a network of devices interacting with different game servers while users play video games.
Unlike black-box approaches, our method explicitly links statistical features to packet-level protocol states. It is structured into four phases: first, \textbf{Network Traffic Monitoring} (Section \ref{sec:method_nta}) gathers data through non-intrusive monitoring of a gaming network; second, \textbf{Feature Extraction} (Section \ref{sec:method_fe}) transforms raw packets into structured windows; third, \textbf{State Characterization} (Section \ref{sec:method_sc}) maps these windows to distinct protocol states; and finally, \textbf{Network Traffic Modeling} (Section \ref{sec:method_ntm}) constructs the interpretable Petri nets for each state.

\subsection{Network Traffic Monitoring}
\label{sec:method_nta}
Online gaming involves multiple users who interact with each other through game servers. Hence, the overall system can be modeled as a bipartite graph $\mathcal{G}(D,Sr,Ar)$, where $D=\{d_1,\dots,d_m\}$ is the set of $m$ devices used by the users, $Sr=\{sr_1,\dots,sr_o\}$ is the set of $o$ game servers, and $Ar$ is the set of network arcs connecting devices with game servers. Each device $d\in D$ can communicate with a game server $sr\in Sr$, hence there are at most $m\times o$ arcs connecting devices and game servers. The network data can be modeled as $\mathcal{P}=(\mu, \rho, \pi)^\alpha$, where $\mu\in\mathcal{C}^a\times \mathbb{R}^b$ indicates the metadata split into $a$ categorical features and $b$ numerical features, $\rho\in\{0,1\}^*$ indicates the (binary) payload, $\pi\in\Pi$ indicates the protocol among the universe of protocols $\Pi$, and $\alpha\in\mathbb{N}$ indicates the number of packets. Each device $d\in D$ generates inbound and outbound network traffic, leading to incoming and outgoing packets that can be collected through non-invasive \textbf{network monitoring}. However, these data are raw, e.g., PCAP files, and unsuitable for the application of process mining techniques. Hence, the subsequent phases will lay out the pre-processing steps required to handle the device data $\mathcal{P}_d$ and ``unleash'' process mining techniques.

\subsection{Feature Extraction}
\label{sec:method_fe}
In the feature extraction phase, we aim to filter $\mathcal{P}_d$ to 1) identify packets related to a specific network protocol, and 2) extract structured data that can be handled in the next phase to characterize different network states. First, \textbf{protocol parsing} 
involves selecting a specific network protocol and extracting protocol-wise network data. We will use the TCP protocol as a running example, as it supports message passing for higher-level IoT protocols such as MQTT. However, the framework also suits other transport-level and application-level protocols due to its general formulation. The network data can be parsed to isolate TCP traffic and obtain $\mathcal{P}_{d,T}$, where $T$ indicates the TCP protocol. $\mathcal{P}_{d,T}$ data could include different information, such as the source IP and port, the TCP flag, and the payload size of the TCP packets. As protocol data are isolated, we proceed to extract synthetic features useful to characterize the traffic. This is done through feature extraction by \textbf{windowing} $W:(\mathcal{C}^a\times \mathbb{R}^b\times \{0,1\}^*\times \Pi)^\alpha\rightarrow \mathbb{R}^{\frac{\alpha}{WL}\times f}$. This function applies a sliding window of length WL and extracts $\frac{\alpha}{WL}$ windows with $f$ features from the protocol data, i.e., $W(\mathcal{P}_{d,T})=\mathcal{P}_{d,T,f}$. These features may include, e.g., the number of specific TCP flags (ACK, SYN, FIN, etc.) and the average payload size of the windows scanned in the network data. The window length used to obtain the $\frac{\alpha}{WL}$ windows determines the length of the network traces that will be considered in the network traffic modeling step. It is worth noting that the $WL$ parameter determines the specificity of the packet-level analysis. Longer windows may be able to capture longer TCP event patterns. However, this also heavily influences the method results due to the underfitting effect typical of process mining techniques, which will be detailed in the experimentation of Section \ref{sec:experiments}.

\subsection{State Characterization}
\label{sec:method_sc}
This phase aims to identify different network states from $\mathcal{P}_{d,T,f}$. This allows separating different types of TCP traffic flows and opens the opportunity to use process discovery in the subsequent phase. On account of the unavailability of labeled data flows, we integrate an unsupervised process through the application of a \textbf{clustering} function $\mathcal{S}$ to the packet statistics. Let $S=\{s_i\in\mathbb{R}^f:i\in\mathbb{N}\}$ be a set of centroids in the $\mathbb{R}^f$ space. The cardinality $n$ of $S$ determines the dimension of the state space. The state space influences the specificity of the packet-level analysis; in particular, fewer states may over-generalize the TCP sequence patterns.
$\mathcal{S}:\mathbb{R}^{\frac{\alpha}{WL}\times f}\rightarrow S^\frac{\alpha}{WL}$ associates a state to each window in $\mathcal{P}_{d,T,f}$. Once the set of states $S_{\mathcal{P}}$ associated with the windows is obtained, these need to be connected with the original protocol data in order to subsequently extract TCP events. To this aim, we implement a \textbf{protocol data alignment} function that associates each state found in the windows with the TCP protocol data. This function $\mathcal{A}:(\mathcal{C}^a\times \mathbb{R}^b\times \{0,1\}^*\times \Pi)^\alpha\times S^\frac{\alpha}{WL}\rightarrow (\mathcal{C}^a\times \mathbb{R}^b\times \{0,1\}^*\times \Pi\times S)^\alpha$ is such that $\mathcal{A}(\mathcal{P}_{d,T}, S_{\mathcal{P}})$ associates a state with each original TCP packet, leading to state-wise protocol data $\mathcal{P}_{d,T,S}\in (\mathcal{C}^a\times \mathbb{R}^b\times \{0,1\}^*\times \Pi \times S)^\alpha$. In this paper, we only consider non-overlapping windows. Therefore, given a window of $WL$ packets to which a state $s\in S$ is assigned, the state is simply replicated for each packet of the window. In conclusion, this process led to the association of a state with each window of TCP packets in the original network data.

\subsection{Network Traffic Modeling}
\label{sec:method_ntm}
Once state-wise protocol data are obtained, different event logs containing the network traces of each state are extracted. Let us define an event log as a set of $k$ traces $L=\{\sigma_1,\sigma_2,\dots,\sigma_k\}\in\mathcal{B}(\Sigma^*)$, where $\Sigma$ denotes the set of TCP events, $\Sigma^*$ the universe of traces that can be built with $\Sigma$, and $\mathcal{B}(\Sigma^*)$ the set of bags over $\Sigma^*$.
To apply this definition, \textbf{event log extraction} $\mathcal{E}:(\mathcal{C}^a\times \mathbb{R}^b\times \{0,1\}^*\times \Pi \times S)^\alpha\rightarrow (\mathcal{B}(\Sigma^*))^{|S|}$ splits the state-wise packet data into $n$ partitions and builds an event log for each state. Specifically, let us denote $\mathcal{P}_{d,T,s_i}$ the state-wise protocol data of $d$ for the $i$-th state. $\mathcal{P}_{d,T,s_i}$ is further split into different subsequences according to the window length used during windowing. Once the different subsequences are obtained, each is converted into a trace collecting the events within the subsequence. The set of traces built from $\mathcal{P}_{d,T,s_i}$ results in the event log $L_{d,T,s_i}$.

Next, \textbf{process discovery} involves finding the relationships between the events of $L_{d,T,s_i}$. It is worth noting that, as remarked in the challenges outlined in Section \ref{sec:intro}, the noisy and interleaved nature of network data may lead to complex and underfitting models. 
One of the most popular formalisms employed in the process mining community is the Petri net \cite{aalst2022pmhb}, which is a bipartite graph consisting of places and transitions. Transitions are mapped to either events found in the event log or to the so-called $\tau$-labeled silent events, which are included to account for unobserved behavior or to introduce, e.g., loop patterns. The nodes of the Petri net are linked through arcs. The overall structure describes the possible control flows that the Petri net allows. The interpretability of the resulting Petri net can be evaluated through the simplicity metric, which depends on the structural properties of the Petri net and can be measured through the arc degree \cite{aalst2022pmhb}.
In addition, the quality of the Petri net can be evaluated through alignment-based conformance checking, which is the state-of-the-art variant of process mining algorithms to evaluate the alignment of an event log with the Petri net \cite{vitale2025cfadpm}.
In conclusion, by applying a process discovery algorithm $\gamma$ to $L_{d,T,s_i}$, we obtain a Petri net $N_{d,T,s_i}$ that models the $i$-th state of the device traffic, i.e., $\gamma(L_{d,T,s_i})=N_{d,T,s_i}$. 

\section{Evaluation}
\label{sec:experiments}
Our experimentation aims at evaluating the ability of our method to 1) coherently capture interpretable behavioral models of network traffic across different devices, and 2) classify network traffic data through these models.
In the following, we detail the UPSIDE case study, the application of the method's phases, the experimental results, and the modeling analysis.

\begin{figure*}
    \centering
    \includegraphics[width=0.9\linewidth]{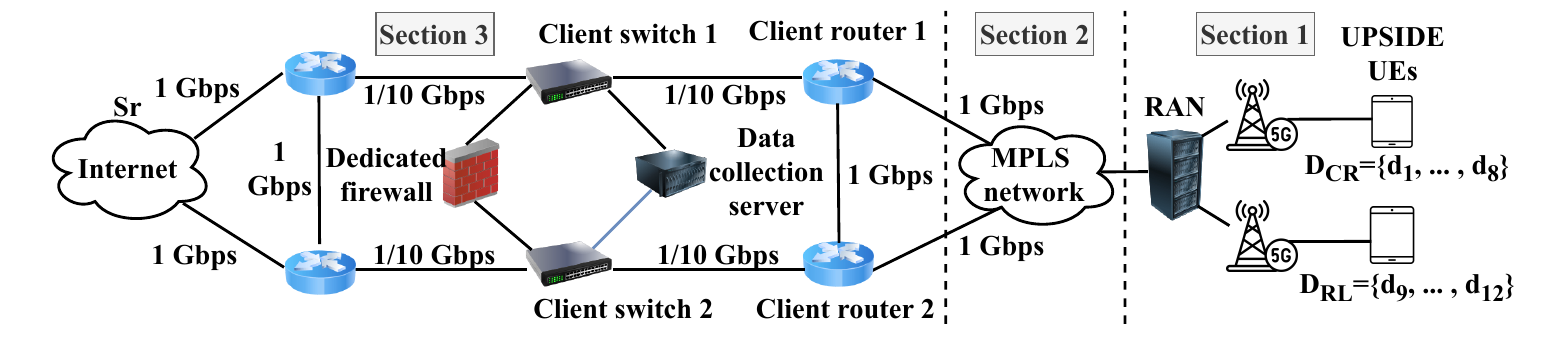}
    \caption{Gaming event network of the UPSIDE case study}
    \label{fig:server_farm}
\end{figure*}
\subsection{The UPSIDE Case Study}
The UPSIDE case study\footnote{https://progettoupside.it/} involves network traffic collected during the UPSIDE gaming event, which featured several parallel sessions of various games \cite{upsideAina2025}. 
The featured games at the event that we analyze are Clash Royale (CR), a real-time strategy game, and Rocket League (RL), an action game.
The gaming event network of the UPSIDE case study is composed of three sections, depicted in Fig. \ref{DATASET_INFO}. The first section consists of two 5G local networks, which were accessed by the event-participating IoT devices using different 5G antennas. Among those are 8 devices $D_{CR}=\{d_1,\dots,d_8\}$ that played CR and 4 devices $D_{RL}=\{d_9,\dots,d_{12}\}$ that played RL. Please note that although these IoT devices are identified by a local IP address, we will refer to the labels in $D_{CR}$ and $D_{RL}$. The second section is an intermediate MPLS network used to route the traffic to the partner ISP's datacenter. The third section is where the traffic is collected and routed to the Internet, where the game servers $Sr$ are located. The data center in this section contains network devices dedicated to the event, such as two sets of routers, one routing from and to the internet and the other for the event. Also, the network switches with traffic mirroring capabilities allowed us to collect the traffic without degrading the network's performance.
Table \ref{DATASET_INFO} summarizes key per-device dataset information: number of TCP packets, number of unique TCP flows, and average TCP flow length.
In the following, we demonstrate the application of the method proposed in Section \ref{sec:method} to a generic UE $d$. The source code implementing the proposed method is available online on GitHub
\footnote{https://github.com/francescovitale/pm\_video\_game\_traffic\_analysis}. Next, we perform two different experiments. The first aims to show the modeling ability and interpretability of the method. The second demonstrates the classification capabilities of the method.

\begin{table}[!t]
\centering
\caption{CR and RL network data information in terms of number of TCP packets, number of TCP flows, and average TCP flow length in terms of TCP packets.}
\label{DATASET_INFO}
\begin{tabular}{lllll}
\hline
\textbf{Game}                                                                   & \textbf{Dev.} & \textbf{\#Packets} & \textbf{\#Flows} & \textbf{Flow len.} \\ \hline
\multirow{8}{*}{\begin{tabular}[c]{@{}l@{}}Clash\\ Royale\\ $D_{CR}$\end{tabular}}  & $d_1$         & 227948             & 2971             & 76                       \\
                                                                                & $d_2$         & 180979             & 2790             & 64                       \\
                                                                                & $d_3$         & 80854              & 1506             & 53                       \\
                                                                                & $d_4$         & 185068             & 2216             & 83                       \\
                                                                                & $d_5$         & 72405              & 1677             & 43                       \\
                                                                                & $d_6$         & 143131             & 2359             & 60                       \\
                                                                                & $d_7$         & 251018             & 2730             & 91                       \\
                                                                                & $d_8$         & 185197             & 3073             & 60                       \\ \hline
\multirow{4}{*}{\begin{tabular}[c]{@{}l@{}}Rocket\\ League\\ $D_{RL}$\end{tabular}} & $d_9$         & 115284             & 2487             & 46                       \\
                                                                                & $d_{10}$      & 39605              & 1034             & 38                       \\
                                                                                & $d_{11}$      & 82148              & 1973             & 41                       \\
                                                                                & $d_{12}$      & 39701              & 1066             & 37                       \\ \hline
\end{tabular}%
\end{table}

\subsection{Method application to UPSIDE}
\label{sec:method_application}

First, network traffic monitoring was performed during the two days of the event, twice each day. The collected network data of each device $d$ reflect this characteristic by being divided into four separate PCAP files, one for each session.

During feature extraction, the PCAP files are parsed to extract TCP traffic. The set of features $\mu$ of the resulting TCP packet data $\mathcal{P}_{d,T}$ of a given device $d$ are the \textit{timestamp}, \textit{direction} (client-to-server/server-to-client), \textit{source\_ip}, \textit{source\_port}, \textit{destination\_ip}, \textit{destination\_port}, \textit{session\_number}, \textit{tcp\_flag}, and \textit{payload\_size}. Next, non-overlapped windowing with a window length $l$ is applied to $\mathcal{P}_{d,T}$, extracting the $\beta$ packet statistics $\mathcal{P}_{u,T,f}$ with $f$ features \textit{avg\_payload}, \textit{n\_servers}, \textit{n\_user\_ports}, \textit{n\_ack}, \textit{n\_syn}, \textit{n\_fin}, \textit{n\_psh}, and \textit{n\_rst}.

In the state characterization phase, $\mathcal{P}_{d,T,f}$ is clustered into $n$ states $s_1,\dots,s_n$ through K-means. The alignment of $\mathcal{P}_{d,T,f}$ with $\mathcal{P}_{d,T}$ occurs as described in Section \ref{sec:method}, resulting in state-wise packet data $\mathcal{P}_{d,T,S}$.

Finally, network traffic modeling extracts event logs from $\mathcal{P}_{d,T,S}$. Firstly, $\mathcal{P}_{d,T,S}$ is split into $n$ different segments $\mathcal{P}_{d,T,s_1},\dots,\mathcal{P}_{d,T,s_n}$. For each segment, a further split extracts traces of $l$ packets each. Hence, for each trace, there are $l$ events. An event is the combination of the \textit{direction} and \textit{tcp\_flag}. By building the events of each trace for all segments, the $n$ event logs $L_{d,T,s_1},\dots,L_{d,T,s_n}$ are built. Process discovery is then applied to each event log through the inductive miner \cite{aalst2022pmhb}, resulting in the $n$ Petri nets $N_{d,T,s_1},\dots,N_{d,T,s_n}$.

\begin{figure*}[!t]
\centering
\includegraphics[width=0.8\textwidth]{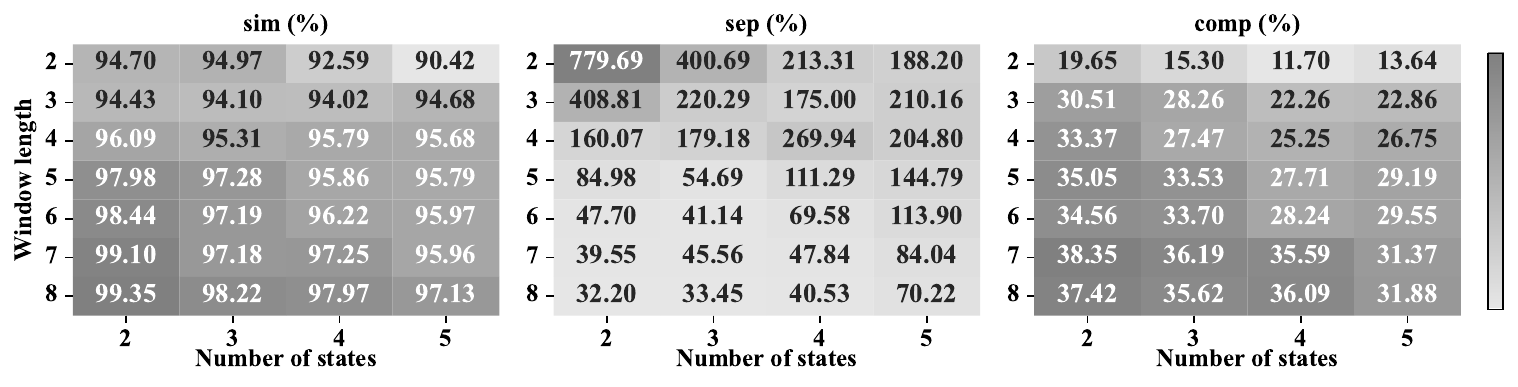}
\caption{Similarity ($sim$), separation ($sep$) and complexity ($comp$) percentages of each set of Petri nets obtained with different window lengths and numbers of states.}
\label{fig:MODELING_HEATMAPS}
\end{figure*}

\subsection{Experiment 1: Modeling CR network data}
This experiment evaluates the modeling capabilities of our method using the RL network data and the fitness metric. The fitness measures how much a Petri net fits the actual behavior of the user, quantifying it through a real value between 0 and 1. However, the fitness metric is sensitive to a cumbersome effect: underfitting. Such an effect is due to the tendency of the inductive miner to generate a model that is \textit{too general}, i.e., too much behavior is allowed, and two different behaviors may both achieve high fitness. To account for this, we use two more metrics: the inter-device similarity ($sim$) and inter-state separation ($sep$) metrics. Let $F_{d_i}(N_{d_j,T,s})$ indicate the fitness between the event log of $d_i$ and the Petri net corresponding to state $s$ of device $d_j$. Let $\bar{F}_{d_i}(s)=\frac{1}{|D_{CR}|}\sum_{d_j\in D_{CR}}(F_{d_i}(N_{d_j,T,s}))$ be the mean fitness obtained by $d_i$ when compared to each $d_j$ of $D_{CR}$ and $\sigma_{d_i}(s)=\sqrt{\frac{1}{|D_{CR}|}\sum_{d_j\in D_{CR}}(F_{d_i}(N_{d_j,T,s})-\bar{F}_{d_i})^2}$ be the corresponding standard deviation. $sim$ and $sep$ are defined as follows:
\begin{flalign*}
    &sim =\frac{1}{|S|}\sum_{s\in S}\frac{1}{|D_{CR}|}\sum_{d_i\in D_{CR}} 1-\frac{\sigma_{d_i}(s)}{\bar{F}_{d_i}(s)}&\\
    &sep = \frac{1}{|S|}\sum_{s\in S}\frac{1}{|D_{CR}|}\sum_{d_i\in D_{CR}}\frac{1}{|S|-1}\sum_{s_j\in S-\{s\}}\frac{F_{d_i}(N_{d_i,T,s})}{F_{d_i}(N_{d_i,T,s_j})} - 1&
\end{flalign*}

The similarity $sim$ is a real number between 0 and 1, where higher values indicate a higher similarity of the Petri nets across the devices. The separation $sep$ is a real number and has two interpretations:
\begin{itemize}
    \item $sep < 0$: the network data of a given device $d_i$ is $|sep|\%$ better fitted by the Petri nets of other devices $d_j,\; j \neq i$, than by the Petri nets of $d_i$ itself;
    \item $sep \geq 0$: the network data of a given device $d_i$ is $sep\%$ better fitted by its own Petri nets than by the Petri nets of other devices $d_j,\; j \neq i$.
\end{itemize}

In addition to evaluating the similarity and separation, it is useful to evaluate the complexity $comp$ of the resulting Petri nets. To do this, we can employ one of the many simplicity metrics proposed by the process mining community: the arc degree of Petri nets, a real number between 0 and 1. Let $arc(\cdot)$ be the arc degree of a Petri net. $comp$ is as follows:
\begin{flalign*}
    &comp=\frac{1}{|S|}\sum_{s\in S}\frac{1}{|D_{CR}|}\sum_{d\in D_{CR}}1-arc(N_{d,T,s})&
\end{flalign*}

We apply the process described in Section \ref{sec:method_application} to each of the CR devices as follows. We take one of the devices out of the set and extract pre-processing parameters from the feature extraction and state characterization steps, including windowing and clustering. The rest of the network data from the other devices is considered the test set and processed with the same parameters, obtaining, for each device, $n$ Petri nets representing states $s_1,\dots,s_n$. Finally, we check the inter-device similarity and inter-state separation with the above formulas using different window lengths and numbers of states.

Fig. \ref{fig:MODELING_HEATMAPS} shows the $sim$, $sep$ and $comp$ results obtained for different window lengths and numbers of states. First, let us analyze these three metrics as the number of states increases. In this case, the $sim$, $sep$ and $comp$ percentages tend to drop. This is particularly evident for a window length equal to 2, where $sim$, $sep$ and $comp$ drop from 94.70\%, 779.69\% and 19.65\% to 90.42\%, 188.20\% and 13.64\%. However, although $sim$ and $sep$ drop, a bigger state space could potentially lead to better classification performances, as the Petri nets may tend to describe more specific behaviors found in the network data. This will be proven true in the next experiment.

\begin{table*}
\centering
\resizebox{\textwidth}{!}{%
\begin{tabular}{llllllllllllllll}
\hline
\textbf{WL} & \multicolumn{15}{l}{\textbf{Number of states}}                                                                                                                                                                                                                                                 \\ \cline{2-16} 
\textbf{}   & \multicolumn{3}{l}{2 states}               & \textbf{} & \multicolumn{3}{l}{3 states}                                                                                      & \textbf{} & \multicolumn{3}{l}{4 states}                & \textbf{} & \multicolumn{3}{l}{5 states}                \\ \cline{2-4} \cline{6-8} \cline{10-12} \cline{14-16} 
\textbf{}   & $CosSim$ (\%)  & AUC (\%)       & $S$ (\%) &           & $CosSim$ (\%)                          & AUC (\%)                               & $S$ (\%)                        &           & $CosSim$ (\%)   & AUC (\%)       & $S$ (\%) &           & $CosSim$ (\%)   & AUC (\%)       & $S$ (\%) \\ \hline
2           & $80.62_{6.03}$ & $68.10_{4.32}$ & $78.97$  &           & $76.91_{4.65}$                         & $72.94_{3.71}$                         & $85.35$                         &           & $38.83_{16.12}$ & $87.46_{6.47}$ & $88.31$  &           & $39.95_{13.13}$ & $87.88_{5.37}$ & $87.51$  \\
3           & $70.76_{3.12}$ & $77.48_{0.45}$ & $70.56$  &           & \cellcolor[HTML]{DFDFDF}$58.14_{9.72}$ & \cellcolor[HTML]{DFDFDF}$88.30_{3.97}$ & \cellcolor[HTML]{DFDFDF}$81.78$ &           & $62.66_{11.92}$ & $81.08_{4.51}$ & $81.95$  &           & $80.42_{8.28}$  & $62.19_{6.33}$ & $77.09$  \\
4           & $79.14_{2.82}$ & $70.05_{2.08}$ & $68.90$  &           & $71.22_{3.05}$                         & $76.92_{0.95}$                         & $78.04$                         &           & $81.04_{3.12}$  & $75.03_{0.84}$ & $75.23$  &           & $78.42_{3.63}$  & $76.25_{1.31}$ & $72.49$  \\
5           & $72.24_{2.38}$ & $78.54_{0.77}$ & $72.22$  &           & $67.45_{2.64}$                         & $80.46_{2.18}$                         & $74.68$                         &           & $67.98_{1.56}$  & $79.07_{3.27}$ & $71.52$  &           & $50.55_{8.60}$  & $85.12_{3.15}$ & $74.14$  \\
6           & $96.19_{3.30}$ & $59.39_{6.39}$ & $63.96$  &           & $83.90_{2.36}$                         & $70.09_{2.87}$                         & $64.72$                         &           & $83.89_{5.35}$  & $71.76_{4.99}$ & $65.39$  &           & $78.10_{6.50}$  & $80.13_{2.64}$ & $71.50$  \\
7           & $97.92_{4.24}$ & $55.81_{8.41}$ & $61.19$  &           & $84.79_{4.41}$                         & $76.65_{5.73}$                         & $69.80$                         &           & $75.14_{2.55}$  & $81.31_{1.84}$ & $70.61$  &           & $86.75_{4.74}$  & $78.30_{4.96}$ & $70.02$  \\
8           & $99.98_{0.02}$ & $52.25_{1.02}$ & $64.10$  &           & $90.34_{3.82}$                         & $64.04_{2.92}$                         & $62.99$                         &           & $90.68_{1.85}$  & $63.14_{1.32}$ & $63.88$  &           & $91.08_{2.37}$  & $62.75_{1.39}$ & $63.17$  \\ \hline
\multicolumn{16}{l}{\textbf{Kruskal-Wallis Test p-values:} AUC (WL: $p< 0.001$, States: $p< 0.001$); CosSim (WL: $p< 0.001$, States: $p=0.007$)}                                                                                                                                                             \\ \hline
\end{tabular}%
}
\caption{Intersection ($I$) cosine similarity ($CosSim$) and AUC per window length (WL)-number of states configuration. The grey cell highlights the configuration with the least similarity between P$_{CR}$ and P$_{RL}$ and the best AUC.}
\label{table:exp_2}
\end{table*}
As regards increasing window length values, the $sim$ percentage increases, the $sep$ percentage decreases and the $comp$ percentage increases. For example, for a number of states equal to 2, the $sim$ percentage goes from 94.70\% to 99.35\% while $sep$ drops to 32.20\% and $comp$ peaks to 37.42\%. This can be due to the underfitting phenomenon, in which heterogeneous behaviors are squeezed into complex Petri nets, fitting behaviors that should be otherwise differentiated from each other. The underfitting effect of increasing the window length is partially mitigated by increasing the number of states. 

As a final remark, the Petri nets obtained with window length and number of states equal to 2 appear to be the best ones, as $sim$ and $sep$ percentages are both very high (94.70\% and 779.69\%). However, as it will be shown later, a low number of states always leads to poor classification performances, as a simpler state space makes shallower generalizations, despite the high $sep$ percentage.

\subsection{Experiment 2: Classification of network data}
\begin{figure}[!t]
\centering
\includegraphics[width=0.65\columnwidth]{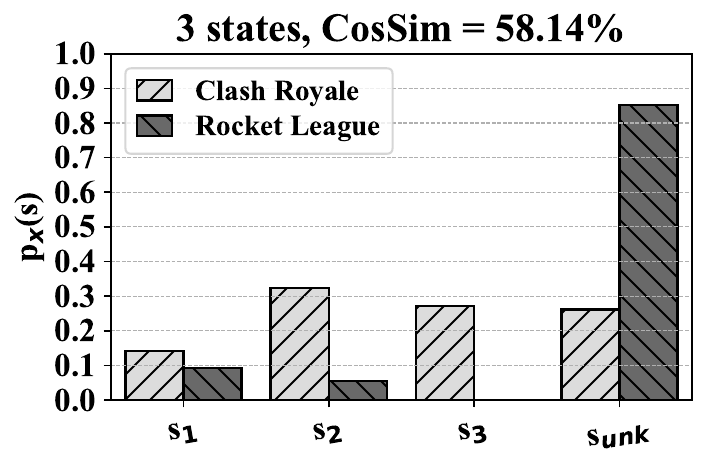}
\caption{The probability mass functions of CR ($P_{CR}$) and RL ($P_{RL}$) for 3 states with a WL equal to 3.}
\label{fig:PMF_COMPARISON}
\end{figure}

We split the CR devices $d_{1,\dots,8}$ as follows. Device $d_1$ serves as the training device, whose data are used to compute all preprocessing parameters and to build one training Petri net per state. Devices $d_{2,\dots,8}$ are used to construct the validation and CR test sets. The validation set is used to determine the fitness threshold for state assignment by selecting the 50th percentile value. The test set is used for classification using this threshold. To account for randomness, devices are randomly distributed across five runs; in each run, three devices are assigned to the validation set and four to the test set. For each run, the test set is extended with the RL devices $d_{9,\dots,12}$. All validation and test devices are processed only up to event-log extraction, producing one event log per state. For each state, we compute a fitness threshold using the corresponding validation event log. The traces of that log are replayed on the training Petri net, yielding a set of fitness values. The fitness threshold is used to classify the traces of the CR and RL test sets: a new trace with fitness equal to or above this value is classified as positive, otherwise as unknown.

The classification ability of our method is evaluated with the test set. Specifically, given a state, we take the traces of the test CR devices and test RL devices and evaluate the positive and unknown classifications. Based on these classifications, we evaluate the similarity of two probability mass functions that estimate, based on the positive and unknown classifications, the distribution of states in the CR and RL data. Let $P_{x}=\{p_x(s)\in[0,1]:s\in \bar{S}=S\,\cup\,\{s_{unk}\}\}$ be the probability mass function of video game $x\in\{CR,RL\}$, where $p(s)$ is the probability of state $s$ and $s_{unk}$ indicates a state that could not be identified using the state-wise thresholds mentioned above. We measure the similarity of the two PMFs with the cosine similarity $CosSim$ metric:
\begin{flalign*}
    &CosSim=\frac{\langle P_{CR},P_{RL}\rangle}{||P_{CR}||\cdot||P_{RL}||}&
\end{flalign*}
The lower this metric, the easier it is to discriminate the CR traffic from RL traffic. To further demonstrate the ability of classifying different network traffic, we adopt the Area Under the Receiving Operating Curve (AUC), which evaluates the quality of a classifier built based on the number of states classified as unknown. We compute the AUC by splitting the entire network data of devices $d_{5,\dots,12}$ into smaller segments whose length is 1\% of the total number of packets. The classifier's AUC depends on 1) the ability to identify states in the segments of test CR devices --- true negatives --- and 2) the assignment of unknown states to the segments of test RL devices --- true positives. To evaluate the statistical robustness of the results, we perform the non-parametric Kruskal-Wallis test against the results across the five runs, and further contextualize the results by comparing them with other interpretable one-class classification methods: the Isolation Forest (iForest), Histogram-based Outlier Score (HBOS), Z-score, and Copula-Based Outlier Detection (COPOD). These methods are trained to recognize CR windows obtained with our method's phase 2, and tested with the same strategy above, i.e., the ability to identify positives and unknowns within the segments.
\begin{table}
\centering
\caption{AUC percentages achieved by our method (number of states equal to 3) and other interpretable one-class methods for WL equal to 3.}
\begin{tabular}{lllll}\hline
\textbf{Our method} & iForest & HBOS & Z-score & COPOD \\ \hline
$88.30_{3.97}$ & $76.71_{4.04}$ & $78.55_{2.84}$ & $50.67_{5.55}$ & $54.70_{4.82}$ \\ \hline\end{tabular}
\label{table:comparison}
\end{table}

\begin{figure}[t]
\centering
\includegraphics[width=0.95\columnwidth]{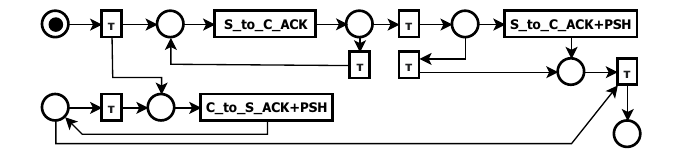}
\caption{The Petri net associated with $s_2$ for WL equal to 3 and 3 states.}
\label{fig:PETRI_NET}
\end{figure}

Table \ref{table:exp_2} reports the $CosSim$, AUC, and simplicity percentages for each window length (WL) and number of states. 
The strongest factor influencing performance is WL. For instance, $\text{WL}=8$ yields high \textit{CosSim} ($99.98\%$) but low AUC ($52.25\%$), an impact supported by $p<0.001$ for both metrics. State-space complexity shows a slightly weaker relation ($p = 0.007$); while using only 2 states never yields the best results, a simpler state space can sometimes outperform a more complex one. The optimal trade-off is achieved at $\text{WL}=3$ with 3 states ($\textit{CosSim}=58.14\%$, $\text{AUC}=88.30\%$), suggesting that WL should not be excessive and complexity must be balanced. Finally, larger WLs tend to simplify the training Petri nets, whereas larger state spaces simplify individual states due to higher fragmentation.
Figure \ref{fig:PETRI_NET} provides further insight into the results achieved by the optimal trade-off, where the comparison of the probability mass functions $P_{CR}$ and $P_{RL}$ outlines that a significant portion of RL data cannot be recognized. 
Finally, Table \ref{table:comparison} compares the performance of our method with the other interpretable one-class methods, outlining its superior classification abilities.

As a final analysis of the interpretability of the method, we show in Figure \ref{fig:PETRI_NET} the Petri net associated with state $s_2$ for window length equal to 3 and 4 states.
The Petri net (the simplest among the four available, selected for conciseness) represents a burst of messages transmitted from the client to the game server (\texttt{C\_to\_S\_ACK+PSH}) carrying the \textbf{PSH} flag, which indicates that the data should be immediately forwarded to the server without buffering. This behavior characterizes a typical communication pattern in CR, where the client continuously transmits numerous small packets to the server. Concurrently, the server issues acknowledgment (\textbf{ACK}) packets, terminating the burst with a final transmission that directly delivers the content to the client (the \texttt{S\_to\_C\_ACK+PSH} transition). It is noteworthy that such a communication pattern was entirely identified through an automated analysis process.

\section{Conclusion}
\label{sec:conclusions}
The growing popularity of online gaming is increasingly attracting research efforts aimed at improving the quality of service provided to users, particularly in the context of dedicated physical events that involve many concurrent users connected simultaneously. 

This paper proposes an unsupervised, process mining-based method for modeling the traffic of network data generated by devices interacting with game servers while diverse video games are played. We applied the method to traffic data of the UPSIDE case study, which involves network data captured from multiple devices while attendees were playing Clash Royale and Rocket League. Results demonstrate that the gaming network behavior can be effectively modeled through states represented as interpretable Petri nets while maintaining a good classification accuracy of different video games (88.30\% AUC), competing with other interpretable one-class classifiers. 

The results highlight three trends: 1) larger window lengths degrade classification; 2) balancing state complexity improves performance regardless of window length; and 3) large window lengths combined with simple state spaces reduce Petri net interpretability. Points 1) and 3) stem from process mining underfitting, which creates shallow generalizations by merging heterogeneous behaviors. Conversely, point 2) reflects the balance between underfitting and overfitting (specialization to specific patterns). Future work will refine the method to reduce behavioral overlap, provide heuristics for selecting optimal Petri nets, and strengthen generalization by analyzing overlapping windows and other IoT protocols.

\bibliographystyle{IEEEtran}
\bibliography{bibliography}

\end{document}